\documentclass[10pt,twocolumn,letterpaper]{article}

\usepackage[pagenumbers]{cvpr} % To force page numbers, e.g. for an arXiv version

\usepackage{tcolorbox}
\usepackage{colortbl}
\usepackage{threeparttable}
\usepackage{pifont} 
\usepackage{multirow}
\usepackage{makecell}
\usepackage{array}
\tcbuselibrary{breakable}
\definecolor{cvprblue}{rgb}{0.21,0.49,0.74}
\usepackage[pagebackref,breaklinks,colorlinks,allcolors=cvprblue]{hyperref}
\definecolor{lightblue}{HTML}{E8F0FE}
\definecolor{blue}{HTML}{013371}
\newcommand{\cc}[0]{\cellcolor{lightblue}}

\def\paperID{*****} % *** Enter the Paper ID here
\def\confName{CVPR}
\def\confYear{2026}
\definecolor{ours}{RGB}{232, 240, 254}
\title{Select Less, Reason More: Prioritizing Evidence Purity for Video Reasoning}

\author{\textbf{Xuchen Li}$^{1,2,3}$\thanks{Equal contribution.} \hspace{9pt} 
\textbf{Xuzhao Li}$^{4*}$\hspace{9pt}
\textbf{Shiyu Hu}$^{4}$\hspace{9pt}
\textbf{Kaiqi Huang}$^{1,2}$\thanks{Corresponding Author.} \hspace{9pt}\\
\textsuperscript{1}CASIA,
\textsuperscript{2}UCAS,
\textsuperscript{3}ZGCA,
\textsuperscript{4}NTU\\
\tt\small s-lxc24@bjzgca.edu.cn, xuzhaoli2001@gmail.com, kaiqi.huang@nlpr.ia.ac.cn
}
\begin{document}
\maketitle
\begin{abstract}
% Long-form video reasoning remains a major challenge for Video Large Language Models (Video LLMs), as static uniform frame sampling leads to information dilution and obscures critical evidence. Furthermore, existing pixel-space video reasoning agents, which are designed to actively interact with the video to acquire new visual information, remain suboptimal due to their lack of rigorous reward mechanisms to enforce evidence purity and their inability to perform temporal information supplementation beyond pre-sampled frames. To address this critical gap, we propose a novel evidence-prioritized adaptive framework built upon our core philosophy: “Select Less, Reason More.” Our core contribution is the evidence-aware reinforcement learning (EARL) framework, which transforms the model into an active interrogator of evidence. EARL is precisely engineered to dynamically select the most relevant frames and, crucially, to perform localized re-sampling around the selected key frames to access fine-grained temporal detail. Strategic policy learning is achieved via a novel multi-component reward system specifically designed to enforce evidence purity through a relevance reward, an IoU-constrained correctness reward, and a dynamic adjustment mechanism for stable convergence. Extensive experiments on five demanding video reasoning benchmarks demonstrate that our EARL-trained model achieves new state-of-the-art among open-source Video LLMs, simultaneously learning an effective and high-purity visual evidence selection policy.
Long-form video reasoning remains a major challenge for Video Large Language Models (Video LLMs), as static uniform frame sampling leads to information dilution and obscures critical evidence. Furthermore, existing pixel-space video reasoning agents, which are designed to actively interact with the video to acquire new visual information, remain suboptimal due to their lack of rigorous reward mechanisms to enforce evidence purity and their inability to perform temporal information supplementation beyond pre-sampled frames. To address this critical gap, we propose a novel evidence-prioritized adaptive framework built upon our core philosophy: “Select Less, Reason More.” Our core contribution is the evidence-aware reinforcement learning (EARL) framework, which transforms the model into an active interrogator of evidence. EARL is precisely engineered to dynamically select the most relevant frames and, crucially, to perform localized re-sampling around the selected key frames to access fine-grained temporal detail. Extensive experiments on five demanding video reasoning benchmarks demonstrate that our EARL-trained model achieves new state-of-the-art among open-source Video LLMs, simultaneously learning an effective and high-purity visual evidence selection policy. Impressively, our 7B model achieves 59.8\% on LongVideoBench, 69.0\% on MVBench and 64.9\% on VideoMME. These results highlight the importance of prioritizing evidence purity and the effectiveness of our framework.
\end{abstract}
\section{Introduction}
\begin{figure}[t!]
  \centering
   \includegraphics[width=1\linewidth]{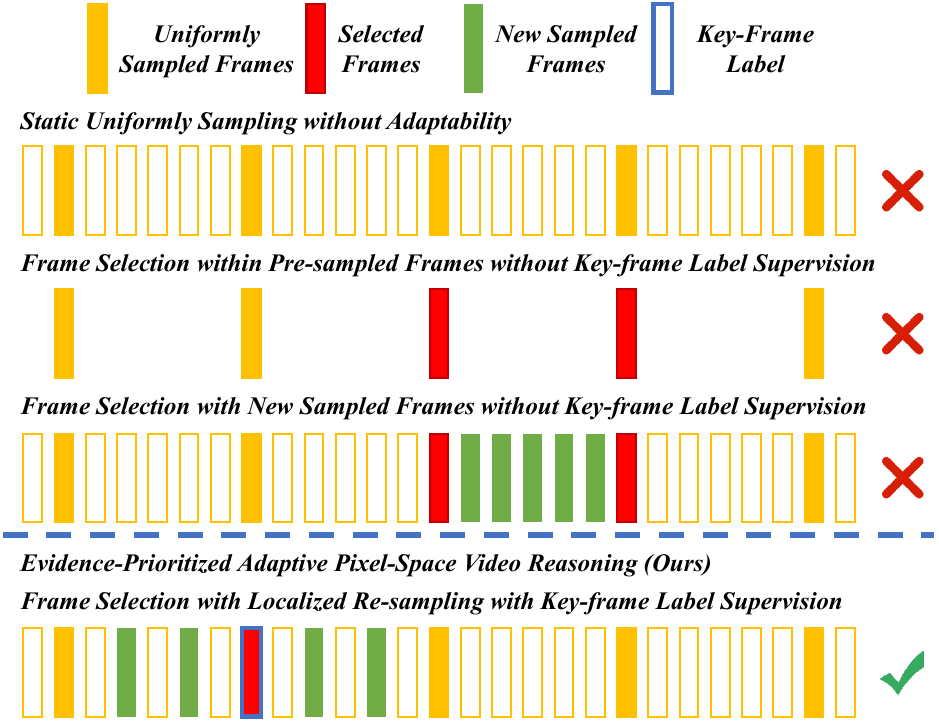}
   \caption{The core motivation and mechanism of our evidence-prioritized adaptive pixel-space video reasoning framework. Existing approaches are limited by three factors ($\times$): 1) Static uniformly sampling dilutes the context with redundant frames; 2) Frame selection within pre-sampled frames restricts access to necessary fine-grained temporal detail; and 3) Selection with new sampled frames without key-frame label supervision fails to enforce evidence purity, potentially leading to sampling in irrelevant areas. Our proposed method ($\checkmark$) overcomes these limitations by integrating frame selection with localized re-sampling to acquire fine-grained temporal detail, and applying key-frame label supervision (via the IoU-based reward) to ensure high evidence purity.}
   \label{fig:motivation}
   \vspace{-15pt} 
\end{figure}

Video Large Language Models (Video LLMs) have made substantial progress in video understanding, primarily owing to their seamless integration of robust visual feature extraction with the advanced capabilities of LLMs \cite{qwen25vl,qwen2vl,gpt4o,internvl3,internvl3.5,deepseekvl,gemini,gemini2.0flash,gemini2.5pro,dtllmvlt}. However, their application in long-form video reasoning \cite{vot,videor1} faces considerable limitations stemming from the video's intrinsic characteristics, which present complex long-range temporal and spatial relationships \cite{mgit,causalstep,li2024visual,li2024dtvlt,fiova}. The prevalent approach of uniform frame sampling fails to address this challenge, as it often dilutes the limited visual context window with redundant information, obscuring the crucial evidence required for precise \cite{wei2024visual,wang2024retake,memvlt}, causality-based decision-making.

To alleviate this, some researchers have explored frame selection methods \cite{mahamad2025key,park2024too,tan2024koala} to pre-determine key frames, often by leveraging external tools like text-visual similarity metrics between the query and the frames. While these methods enhance reasoning by focusing on static, pre-selected visual evidence, they fundamentally operate within the domain of textual-space video reasoning \cite{pixelreasoner}. They treat the visual input as a fixed starting condition, lacking the crucial ability to allow the model to dynamically request and acquire further visual information based on knowledge gaps identified during the reasoning process \cite{lookless,verifybench,darter}.

More recently, the field has progressed toward pixel-space video reasoning, where models are empowered to actively interact with the video and obtain necessary information \cite{pixelreasoner}. These approaches generally fall into two categories: multi-agent Video LLMs \cite{videorag1,videomind} and end-to-end agent Video LLMs \cite{pixelreasoner,framethinker,videomtr,vital}. Approaches like VideoRAG \cite{videorag} exemplify the multi-agent paradigm. Its core innovation lies in its dual-channel architecture that uses an external knowledge component to capture cross-video semantic relationships, which is then integrated with the LLM for generation. This reliance on cooperative but decoupled external components limits the possibility of a unified, end-to-end optimization of the entire reasoning and evidence selection policy. Furthermore, existing end-to-end agent methods, such as Pixel Reasoner \cite{pixelreasoner}, VITAL \cite{vital}, and FrameMind \cite{framethinker}, utilize reinforcement learning (RL) training to enable proactive, tool-harnessing interaction with the video. While methods like VITAL and FrameMind advance the field by allowing the model to learn to select frames within a given video interval—thereby obtaining new information during the reasoning process—they share a critical limitation. Specifically, all these existing end-to-end approaches supervise only the coarse actions without rigorously rewarding whether the selected visual contents genuinely contribute to answering the question or enforcing evidence purity. Moreover, methods like Pixel Reasoner restrict selection solely to the pre-sampled frames, failing to provide the model with a mechanism to access the finer temporal granularity necessary for accurate reasoning. This dual failure—the lack of evidence purity rewards and, in some cases, the inability to access fine-grained temporal detail—is the critical gap our work aims to address.

This necessity drives the development of a unified, adaptive strategy: a framework intelligently capable of ensuring evidence purity by commanding the model to select only the most relevant frames (to minimize contextual distraction), thereby enabling the model to reason more with a cleaner and higher-quality context. Furthermore, to address the limitation of only selecting from pre-sampled inputs, this strategy must incorporate a mechanism for temporal refinement that performs localized re-sampling around the currently selected key frames to access the finer granularity needed for accurate decision-making. This principle forms the foundation of our core philosophy: Select Less, Reason More.

To achieve this adaptive capability, we propose a framework for evidence-prioritized adaptive pixel-space video reasoning, where the selection of frames itself constitutes the key reasoning step in the pixel domain. Specifically, our method is designed to dynamically determine which sparse frames are critical for the answer, and based on the selected key frames, perform localized re-sampling to obtain the necessary temporal details to enrich the visual context. Our comprehensive training pipeline initiates with operation-aware supervised fine-tuning (SFT), providing the baseline competence for multi-step, tool-augmented reasoning. Crucially, we then introduce a novel evidence-aware reinforcement learning (EARL) framework, dedicated to transforming this initial, imitation-based competence into a refined, high-accuracy adaptive strategy. The EARL framework is guided by a multi-component reward system specifically engineered to enforce evidence frame purity. This system includes the relevance reward, which actively promotes the “Select Less” objective by applying a IoU based frame selections; the correctness reward with IoU constraint, which enforces evidence purity and requires correct answers to be derived from visually relevant frames; and a dynamic adjustment mechanism, which guarantees stable convergence by dynamically balancing the training focus between answer correctness and long-term selection.

% Extensive experiments on five video reasoning benchmarks unequivocally demonstrate the effectiveness of our approach. Our frames-aware adaptive method establishes a new state-of-the-art among open-source Video LLMs, consistently outperforming baselines. Furthermore, our detailed ablation studies confirm that the FARL framework and each component of its reward system are individually indispensable for achieving superior reasoning accuracy.

Extensive experiments on five video reasoning benchmarks unequivocally demonstrate the effectiveness of our approach. Our evidence-prioritized adaptive method achieves 59.8\% on LongVideoBench \cite{longvideobench} and 69.0\% on MVBench \cite{mvbench}, establishing a new state-of-the-art among open-source Video LLMs. Ablation studies further confirm that the EARL framework and each component of its reward system are indispensable for achieving superior accuracy.

The contributions of this paper can be summarized in the following three aspects:
\begin{itemize}
    \item We propose a novel framework for evidence-prioritized adaptive pixel-space video reasoning, providing a unified strategy to actively address the challenges of information dilution and temporal redundancy in long-form videos.
    \item We introduce the evidence-aware reinforcement learning (EARL) framework, guided by a novel multi-component reward system specifically engineered to enforce evidence purity and strategically manage the selection of visual context.
    \item Our method achieves superior performance across challenging video reasoning benchmarks, demonstrating state-of-the-art accuracy while learning a high-purity visual evidence selection policy.
\end{itemize}
\section{Related Work}

\begin{figure*}[t!]
   \centering
   \includegraphics[width=0.9\linewidth]{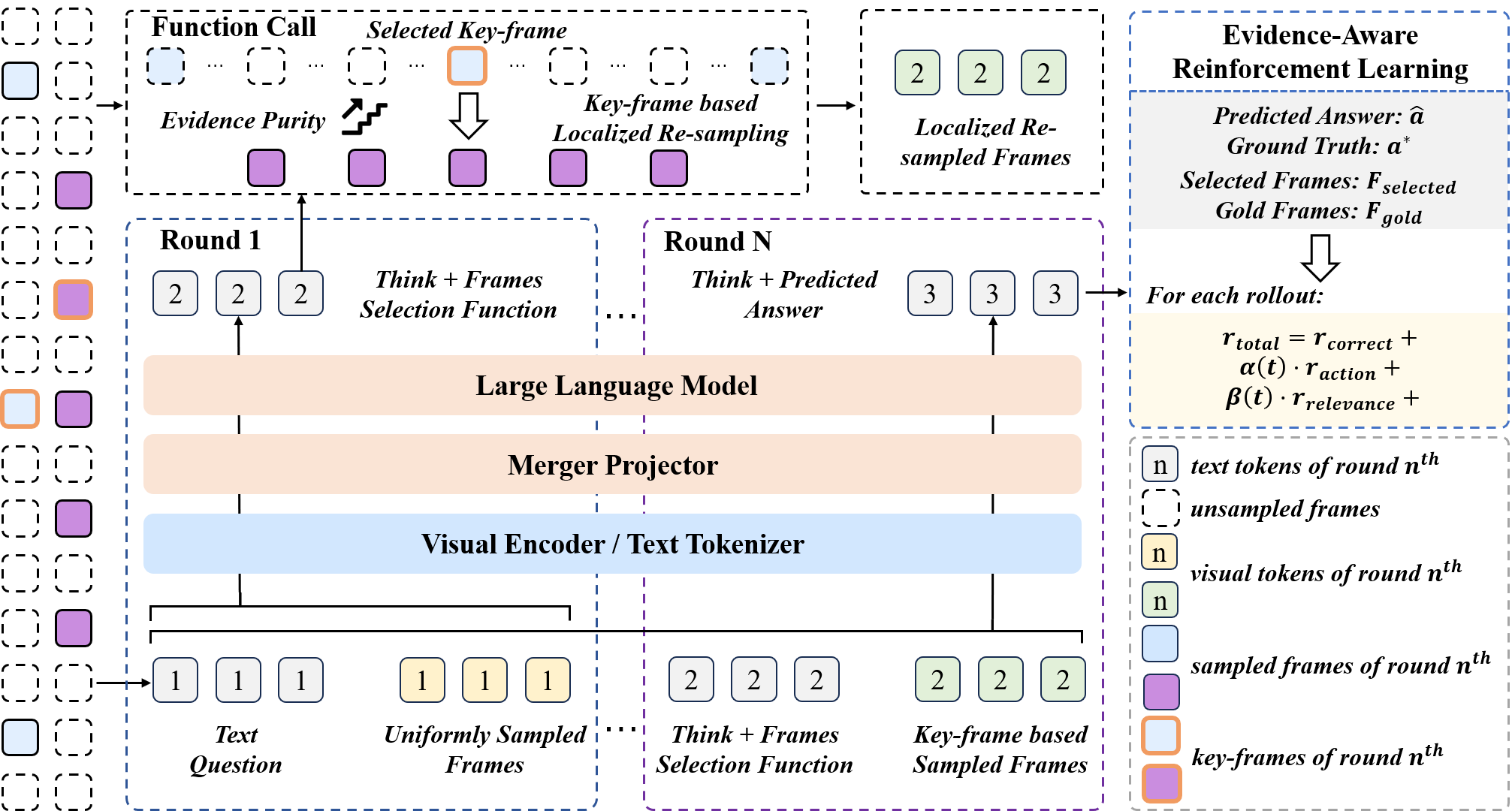}
   \vspace{-5pt}
   \caption{Overview of the Evidence-Aware Reinforcement Learning (EARL) framework. In the multi-round generation process, the model can attend to select frames adaptively and integrate the result of key-frame based localized re-sampling to form a multimodal CoT.}
   \label{fig:method}
   \vspace{-15pt}
\end{figure*}

% \subsection{Video Large Language Models (Video LLMs)}
% The integration of vision encoders and large language models has significantly advanced Video LLMs. Recent influential systems, often built upon paradigms like RLVR, have demonstrated substantial performance gains. However, a critical architectural limitation persists: the prevalent reliance on a passive, uniform frame sampling strategy. This fixed approach, which samples at a constant rate, struggles to scale efficiently to long videos due to the rapid exhaustion of the visual token budget. While token compression, keyframe selection, and long-context finetuning manage this cost, they often operate under static presets, lacking a dynamic solution for coordinating visual bandwidth. Our work directly challenges this fixed-sampling paradigm by proposing a dynamic, frames-aware scheme that actively and intelligently selects a necessary subset of frames, ensuring efficiency without compromising temporal coverage.

\subsection{Textual-space Video Reasoning}
Textual-space reasoning methods \cite{videor1,tinyllavavideor1} focus on enhancing the Video LLM's cognitive process after a fixed visual context is provided. The visual input is treated as a static starting condition, where subsequent complex inference relies heavily on the quality of the textual Chain-of-Thought (CoT) \cite{cot} generated by the LLM. Research in this area often emphasizes post-training \cite{deepseekr1,videorl} to inject structured signals—such as spatio-temporal alignment \cite{fei2024enhancing,li2024texts}—into the reasoning trajectory. While this enhances the model's ability to handle causality and logical flow within the given context, these methods are intrinsically limited by the passive input they receive; they cannot dynamically request new visual information or refine the existing context if the initial, uniformly sampled frames are insufficient or misleading \cite{shang2024traveler,ijcai2025p908}. This static nature prevents the model from resolving ambiguity in information-sparse regions. Our approach directly addresses this limitation by transforming the Video LLM into an active interrogator of evidence, enabling it to dynamically control and purify its visual input.

\subsection{Pixel-space Video Reasoning}
This line of research, often encapsulated by the “Thinking with Images” paradigm \cite{thinking,cao2025large}, addresses the limitations of fixed visual input by empowering the model to actively interrogate visual content through iterative querying. This domain \cite{simpleo3,thyme,revpt,chainoffocus} includes agent-based systems leveraging tools (like indexing) and methods using intra-frame operations (like zoom). Influential works like DeepEyes \cite{deepeyes} utilize RL to incentivize the autonomous use of these tools, treating the query as an intermediate reasoning step. While crucial for enhancing perceptual fidelity, their primary limitation is the lack of a fully autonomous, strategic policy across the entire video; they rely on fixed workflows or remain restricted to single-frame operations. Some recent work \cite{framethinker,vital,videomtr} (e.g., Pixel Reasoner \cite{pixelreasoner}) incorporated video frame selection into end-to-end training, these efforts reinforce only the coarse selection action without incorporating evidence-aware finesse. Our framework proposes a fundamental extension: we introduce the EARL framework to learn an end-to-end policy where evidence-aware adaptive selection is the core pixel-space video reasoning step.
\section{Problem Formulation}
\label{sec:problem}
Video reasoning tasks require models to extract relevant information from long sequences of frames \cite{wang2025multimodal,tang2025video,sun2025survey}. Some queries can be answered by reasoning over general visual patterns, while others depend critically on specific frames that contain temporal or spatial cues. In this work, we focus on \emph{evidence-prioritized adaptive pixel-space video reasoning}, where the model must dynamically select the minimal, yet sufficient, set of frames in order to maximize answer accuracy and evidence purity—the core goal of our “Select Less, Reason More” philosophy.

% Let a video \( V = \{v_1, \dots, v_T\} \) and a question \( Q \) form a query \( \mathbf{x} = [V, Q] \). The model generates a reasoning trajectory \( \mathbf{y} = [y_1, \dots, y_n, \hat{a}] \), where each \( y_t \) corresponds to either a textual reasoning step or a frame-selection action, and \( \hat{a} \) is the final predicted answer. When a frame-selection action selects frame \( v_t \), the visual features of \( v_t \) are incorporated into the reasoning process as follows:
% \begin{align}
%     y_t &\leftarrow \text{concat}(y_t, f_{\text{frame}}(v_t)),
% \end{align}
% where \( f_{\text{frame}}(v_t) \) represents the visual features extracted from the selected frame.

Let a video \( V = \{v_1, \dots, v_M\} \) and a question \( Q \) form a query \( \mathbf{x} = [V, Q] \). The model then generates a reasoning trajectory \( \mathbf{y} = [y_1, \dots, y_n, \hat{a}] \), where each \( y_t \) corresponds to either a textual reasoning step or a frame selection action and $\hat{a}$ represents the predicted answer of Video LLMs.

The model's frame selection action chooses a set of key frames $F_{\text{select}} \subset V_{\text{current}}$ from the current visual context (uniformly sampled frames of $V$). Upon model selection, the system automatically performs a localized re-sampling operation. The localized re-sampling operation identifies a time interval \(\tau_i\) for each selected key frame, between the key frame and its nearest temporally adjacent frame in the current visual context (i.e., the set of uniformly sampled frames \(V_{\text{current}}\)). Then, a total of \(N_{\text{max}}\) frames are uniformly re-sampled from these interval-defined video clips and distributed across \(\tau_i\). This results in a new set of high-granularity frames \( F_{\text{refine}} \), which then becomes the new visual context, i.e., \( V_{\text{current}} \leftarrow F_{\text{refine}} \).

The visual features of the refined and contextually complete frame set $F_{\text{refine}}$ are incorporated into the current reasoning step, as the model acts upon its choice: $y_t \leftarrow \text{concat}(y_t, f_{\text{frame}}(F_{\text{refine}}))$, where $f_{\text{frame}}(F_{\text{refine}})$ represents the combined visual features extracted from the refined frames set.

The correctness reward ($r_{\text{correct}}$) initially reflects only the binary accuracy of the model's prediction $\hat{a}$ against the ground-truth answer $a^*$:
\begin{align}
    r_{\text{correct}}(\mathbf{x}, \mathbf{y}) =
    \begin{cases}
        1 & \text{if } \hat{a} = a^*, \\
        0 & \text{otherwise.}
    \end{cases}
\end{align}
% For our learning objective, this reward will be further refined to incorporate an evidence purity IoU constraint, ensuring that correct answers are derived from relevant visual inputs.

The evidence-awareness reward ($r_{\text{evidence}}$) is designed to enforce the “Select Less” objective. This score incentivizes the model to select only frames crucial to answering the query, specifically by reducing the selection of redundant or irrelevant visual information to ensure evidence purity.

The overall learning objective combines these two components:
\begin{align}
    \max_{\theta} \mathbb{E}_{\mathbf{x} \sim \mathcal{D},\, \mathbf{y} \sim \pi_\theta(\mathbf{y}|\mathbf{x})} \left[ R(\mathbf{x}, \mathbf{y}) \right],
\end{align}
where the total reward \( R(\mathbf{x}, \mathbf{y}) \) is the sum of the correctness reward and the evidence-awareness reward:
\begin{align}
    R(\mathbf{x}, \mathbf{y}) &= r_{\text{correct}}(\hat{a}, a^*) + \lambda \, r_{\text{evidence}}(\mathbf{x}, \mathbf{y}),
\end{align}
and \( \lambda \) is a hyperparameter that controls the trade-off between answer accuracy and adaptive frame selection. The decomposition and precise formulation of $r_{\text{evidence}}$ and refined $r_{\text{correct}}$ will be elaborated in Section \ref{sec:farl}.

\section{Method}
\label{sec:method}
\subsection{Operation-Aware Supervised Fine-Tuning}
\label{sec:sft}
We begin with an operation-aware supervised training phase on $\mathcal{D}_{\text{SFT}}$, a dataset consisting of question-answer pairs along with their corresponding reasoning steps. These reasoning steps, denoted by the trajectory $\mathbf{y}_i$, include both textual CoT steps and explicit frame selection actions which function as callable tools, guiding the model to identify which frames are essential for answering a given query.
% Unlike conventional methods that rely on external pipelines for frame pre-selection, our approach allows the model to learn to select frames dynamically and autonomously during the reasoning process itself.

The model is trained to minimize the standard cross-entropy loss:
\begin{align}
\mathcal{L}_{\text{SFT}} = - \sum_{(\mathbf{x}_i, \mathbf{y}_i) \in \mathcal{D}_{\text{SFT}}} \log P_\theta(\mathbf{y}_i \mid \mathbf{x}_i),
\end{align}
where $\mathbf{x}_i$ represents the input query, $\mathbf{y}_i$ denotes the ground-truth reasoning trajectory, and $\theta$ are the model parameters.

However, SFT is inherently limited by the quality of its expert data; it cannot effectively distinguish between genuinely necessary frame selection and non-optimal actions present in the reasoning trajectories. This limitation necessitates the subsequent refinement through the RL phase, where the model will learn to optimize its decision-making for accuracy and evidence purity.

\subsection{Evidence-Aware Reinforcement Learning}\label{sec:farl}
The evidence-aware reinforcement learning (EARL) phase is the core mechanism that transforms the model's basic operational capability (learned via SFT) into a precise adaptive reasoning policy. As illustrated in Figure \ref{fig:method}, EARL frames the video reasoning task as a sequential decision-making process where the model iteratively alternates between textual reasoning and frames selection operations. After frames selection operations, the model refines its visual context by dynamically performing localized frame re-sampling. To maintain efficiency, the model is strictly limited to a maximum of two dynamic frame selection operations per prompt. This refinement is guided by a multi-component reward system designed to achieve two goals: maximize final answer accuracy and evidence purity. The process is supported by high-quality key frame annotation, which provides the ground truth of golden frames necessary to accurately supervise the relevance and purity of the model's frame selection decisions.

\subsubsection{Key-frame Annotation}
The frame annotation process follows a hybrid approach to establish the ground truth for relevant visual evidence. Initially, the video frames, their corresponding questions, and answers, is provided to GPT-4o \cite{gpt4o}. With the aid of carefully crafted prompts, GPT-4o generates a preliminary set of key frame indices, denoted as \( F_{\text{key}} \), satisfying a size constraint: $|F_{\text{key}}| \in \{1, 2, \dots, 8\}$.

Subsequently, human annotators review the generated set, verifying and eliminating irrelevant frames to ensure evidence purity. The final annotated frame set, \( F_{\text{gold}} \), is obtained by removing any frames deemed non-contributory by the annotators: $F_{\text{gold}} = F_{\text{key}} \setminus F_{\text{irrelevant}}$, where \( F_{\text{irrelevant}} \) represents the frames identified as irrelevant or non-essential by the human reviewers. This $F_{\text{gold}}$ serves as the ground-truth against which the model's selection quality is judged. The annotation process is designed to capture visual evidence required for the model's two-round frame selection.

\subsubsection{Reward Function Design}
The EARL phase refines the model's selection strategy through a multi-component reward system specifically aimed at promoting the “Select Less, Reason More” philosophy. The reward function consists of three primary components: the action reward (\( r_{\text{action}} \)), the relevance reward (\( r_{\text{relevance}} \)), and the correctness reward (\( r_{\text{correct}} \)).

\textbf{Action Reward.} The action reward \( r_{\text{action}} \) incentivizes the model to actively select frames. This is essential to prevent the model from avoiding frame selections due to uncertainty. A small fixed reward is provided for every frame selection action. It is expressed as:
\begin{align}
r_{\text{action}} =
\begin{cases}
1 & \text{if frames are selected}, \\
0 & \text{otherwise}.
\end{cases}
\end{align}

\textbf{Relevance Reward.} The relevance reward \( r_{\text{relevance}} \) encourages the model to select frames that are crucial for answering the query, directly rewarding the purity of the selected set. It is calculated based on the Intersection over Union ($\text{IoU}$) between the selected frames (\( F_{\text{selected}} \)), and the golden key frames (\( F_{\text{gold}} \)). The IoU, which quantifies the overlap, is defined as:
\begin{align}
\text{IoU} = \frac{|F_{\text{selected}} \cap F_{\text{gold}}|}{|F_{\text{selected}} \cup F_{\text{gold}}|}
\end{align}
% A higher $\text{IoU}$ indicates both a greater intersection with $F_{\text{gold}}$ and a smaller union set (i.e., less redundancy).

The relevance reward is a continuous value directly computed as the $\text{IoU}$:
\begin{align}
r_{\text{relevance}} = \text{IoU}.
\end{align}
This reward ranges from $[0, 1]$ and strictly guides the model toward selecting the smallest, purest set of frames that maximally overlaps with the ground truth.

\textbf{Correctness Reward.} The correctness reward \( r_{\text{correct}} \) links the frame selection quality to the ultimate task objective and enforces evidence purity. The core design of this reward mechanism is as follows: when the model's predicted answer \(\hat{a}\) matches the ground truth answer \(a^*\), differential positive rewards are granted based on the IoU of the selected frames. Specifically, a higher reward is given if the IoU is no less than 0.5, while a reward of 0.5 points is given if the IoU is less than 0.5. This reward is expressed as:
\begin{align}
r_{\text{correct}} =
\begin{cases}
1 & \text{if } \hat{a} = a^* \text{ and IoU} \geq 0.5, \\
0.5 & \text{if } \hat{a} = a^* \text{ but IoU} < 0.5, \\
-1 & \text{if } \hat{a} \neq a^*.
\end{cases}
\end{align}
This structure incentivizes the model to not only produce accurate answers but also ensure that those answers are derived from a high-purity set of visual evidence.

\subsubsection{Dynamic Adjustment of Reward Sensitivity}
To improve the model's learning stability, we introduce a dynamic adjustment mechanism for reward sensitivity, tailored to the different stages of training. The goal is to prioritize the exploration tendency of $r_{\text{action}}$ during the early phases and the purity requirements of $r_{\text{relevance}}$ and $r_{\text{correct}}$ during later stages. Let \( t \) denote the current training iteration, and \( T \) be the total number of iterations. We define the training progress as \( \text{Progress} = \frac{t}{T} \), which represents the percentage of training completed.

In the early stages of training (\( \text{Progress} \leq P \)), the focus is on encouraging the model to explore a wide range of frames and actions. To achieve this, we set a higher action reward scaling factor \( \alpha_{\text{early}} \) and a lower relevance reward scaling factor \( \beta_{\text{early}} \). This encourages the model to experiment with different frame selections without being overly focused on strict selection purity.

% As training progresses (\( \text{Progress} > P \)), the focus shifts to refining the model’s ability to maximize purity. In this phase, we reduce \( \alpha_{\text{late}} \) (the action reward scaling factor) and increase \( \beta_{\text{late}} \) (the relevance reward scaling factor). This guides the model to strictly prioritize the purity and accuracy requirements embedded in the $\text{IoU}$-based rewards.

As training progresses ($\text{Progress} > P$), the focus shifts to refining the model’s ability to maximize purity. In this phase, we reduce the action reward scaling factor from $\alpha_{\text{early}}$ to $\alpha_{\text{late}}$ and increase the relevance reward scaling factor from $\beta_{\text{early}}$ to $\beta_{\text{late}}$. This guides the model to strictly prioritize the purity and accuracy requirements embedded in the $\text{IoU}$-based rewards.

% Thus, the relationship between the early and late stages of training can be summarized as: $\alpha_{\text{early}} > \alpha_{\text{late}} \quad \text{and} \quad \beta_{\text{early}} < \beta_{\text{late}}$. 
The total reward \( r_{\text{total}} \) is a weighted sum of the individual rewards:
\begin{align}
r_{\text{total}} = r_{\text{correct}} + \alpha(t) \cdot r_{\text{action}} + \beta(t) \cdot r_{\text{relevance}},
\end{align}
where $\alpha(t)$ and $\beta(t)$ are dynamically adjusted according to the training progress. Specifically, their values are switched from $\{\alpha_{\text{early}}, \beta_{\text{early}}\}$ to $\{\alpha_{\text{late}}, \beta_{\text{late}}\}$ once the training exceeds a predefined threshold $P$. This ensures a gradual transition from action exploration to refined, high-accuracy selection performance, fulfilling the core principle of “Select Less, Reason More.”
\section{Experiments}
\label{sec:experiments}

\begin{table*}[t!]
    \caption{Performance of models on five video reasoning benchmarks. Results marked with $*$ are reproduced by ourselves.}
    \vspace{-5pt}
    \centering
    \resizebox{\linewidth}{!}{
      \begin{tabular}{lcc|ccccccccc}
      \toprule
        \multirow{2}{*}{\textbf{Models}} & \multirow{2}{*}{\textbf{Size}} & \multirow{2}{*}{\textbf{\#Frames}} &  \multirow{2}{*}{\textbf{MLVU}} & \multicolumn{2}{c}{\centering \textbf{VideoMME (w/o sub)}} & \multirow{2}{*}{ \textbf{LongVideoBench}} & \multirow{2}{*}{ \textbf{LVBench} } & \multirow{2}{*}{ \textbf{MVBench} } \\
        \cline{5-6}
        &&&& Overall & Long & & \\
        \rowcolor{gray!10} Duration & &  & 3$\sim$120 min & 1$\sim$60 min & 30$\sim$60 min & 0$\sim$60 min & 4101 sec & 5$\sim$35 sec \\
        \midrule
        \textbf{\textit{Proprietary Models}} & & \\
        GPT-4V \cite{gpt4v} & - & 1fps &  - & 60.7 & 56.9 & - & - & 43.5 \\
        GPT-4o \cite{gpt4o} & - & 1fps & 66.2 & 77.2 & 72.1 & 66.7 & 34.7 & - \\
        \midrule
        \textbf{\textit{Open-Source Video LLMs}} & & \\
        LLaMA-VID \cite{llamavid} & 7B & 1fps & 33.2 & - & - & - & 23.9 & - \\
        Video-LLaVA \cite{videollava} & 7B & 8 & 47.3 & 40.4 & 38.1 & 39.1 & - & - \\
        ShareGPT4Video \cite{sharegpt4video} & 8B & 16 & 46.4 & 43.6 & 37.9 & 39.7 & - & 51.2 \\
        LLaVA-NeXT-Video \cite{llavanextvideo} & 7B & 32 & - & 46.5 & - & 43.5 & - & - \\
        VideoLLaMA2 \cite{videollama2} & 7B & 32 & 48.5 & 46.6 & 43.8 & - & - &  45.5 \\
        LongVA \cite{longva} & 7B & 128& 56.3 & 54.3 & 47.6 & - & - & - \\
        VideoChat2 \cite{mvbench} & 7B & 16 & 47.9 & 54.6 & 39.2 & - & - & 51.1 \\
        LLaVA-OneVision \cite{llavaov} & 7B & 32 & 64.7 & 58.2 & 46.7 & - & - & 56.7 \\
        Vamba \cite{vamba} & 10B & 1024 & 65.9 & 57.8 & - & 55.9 & 42.1 & 60.4 \\
        VideoChat-T \cite{mvbench} & 7B & 12 & - & 46.3 & 41.9 & - & - & - \\
        Quicksviewer \cite{quicksviewer} & 7B & 1fps & 61.5 & 56.9 & - & - & - & 55.6 \\
        Video-XL \cite{videoxl} & 7B & 256 & 64.9 & 55.5 & - & 50.7 & - & - \\
        LongVILA \cite{longvila} & 7B & 256 & - & 60.1 & 53.0 & 57.1 & - & 67.1 \\
        LongVU \cite{longvu} & 7B & 1fps & 65.4 & 60.6 & 59.5 & - & - & 66.9 \\
        Hour-LLaVA \cite{hourllava} & 7B & 1fps &  - & 63.6 & 55.0 & 60.4 & 45.6 & - \\
        LongVITA-128k \cite{longvita} & 14B & 256 & - & 66.4 & 58.8 & 60.9 & - & 55.4 \\
        Video-R1 \cite{videor1} & 7B & 32 & 45.4 & 59.3 & 50.2 & - & - & 63.9 \\
        \midrule
        \textbf{\textit{Open-Source multi-agent Video LLMs}} & &  \\
        VideoMind \cite{videomind} & 7B & - & 64.4 & 58.2 & 49.2 & 56.3 & 40.8 & 64.6 \\
        Video-RAG \cite{videorag} & 7B & - & 72.4 & 62.1 & 59.8 & 58.7 & - & - \\
        \midrule
        \textbf{\textit{Open-Source End-to-end Agent Video LLMs}} & &  \\
        Video-MTR \cite{videomtr} & 7B & 32 & 48.4 & 59.0 & 51.0 & - & - & - \\
        Pixel Reasoner \cite{pixelreasoner} & 7B & 16 & - & - & - & - & - & 67.8 \\
        VITAL \cite{vital} & 7B & 1024 & - & 64.1 & 54.0 & - & - & - \\
        FrameMind \cite{framethinker} & 7B & 32 & 48.6 & 60.9 & 57.5 & - & - & 64.2 \\
        \midrule
        \textbf{\textit{Ours}} & &  \\
        Qwen2.5-VL* \cite{qwen25vl} & 7B & 32 & 41.6 & 53.6 & 44.7 & 43.2 & 31.6 & 62.6 \\
        \cc \shortstack{\textbf{Ours}} & \cc \shortstack{\textbf{7B}} & \cc \shortstack{\textbf{32}}
        & \cc \shortstack{\textbf{49.3}}
        & \cc \shortstack{\textbf{64.9}}
        & \cc \shortstack{\textbf{57.8}}
        & \cc \shortstack{\textbf{59.8}}
        & \cc  \shortstack{\textbf{46.2}}
        & \cc \shortstack{\textbf{69.0}}\\
        \bottomrule
        \end{tabular}
    }
    \label{tab:main_results}
    \vspace{-10pt}
\end{table*}

\subsection{Setups}
\textbf{Training.} 
We follow Pixel-Reasoner, utilizing its datasets consisting of 3.8k samples for SFT phase and 8.3k samples for RL phase. For the RL phase, we also perform key frame annotation on the dataset to ensure accurate frame selection. The base model used for training is Qwen2.5-VL-7B-Instruct \cite{qwen25vl}, and we leverage Open-R1 \cite{openr1} for the SFT phase and OpenRLHF \cite{openrlhf} for RL training. For the SFT phase, we employ a batch size of 128 and set the learning rate to \(1 \times 10^{-6}\), with a 10\% warm-up period to ensure stable training. In the RL phase, a cosine learning rate decay schedule is applied, starting with a learning rate of \(1 \times 10^{-6}\). The training process in RL involves sampling 256 prompts per batch, with each prompt generating 8 rollouts. To manage the visual context budget during training and inference, all videos are initially uniformly sampled to a maximum of 32 frames. We enforce that the model is strictly limited to a maximum of 2 dynamic frame selection operations per prompt. And each selection operation triggers local, uniform re-sampling of the original video, with the number of newly sampled frames capped at 16 ($N_{\text{max}}=16$). We provide detailed system prompts and training hyperparameters in Appendix A and Appendix B, respectively.

\textbf{Baseline.}
We compare our approach against a diverse set of state-of-the-art video reasoning models, including both general-purpose and agent Video LLMs. We first consider proprietary models, such as GPT-4V \cite{gpt4v} and GPT-4o \cite{gpt4o}, which are strong general-purpose systems capable of multimodal reasoning. In addition, we evaluate open-source video LLMs like Video-LLaVA \cite{videollava}, LLaMA-VID \cite{llamavid}, ShareGPT4Video \cite{sharegpt4video}, LLaVA-NeXT-Video \cite{llavanextvideo}, VideoLLaMA2 \cite{videollama2}, LongVA \cite{longva}, VideoChat2 \cite{mvbench}, LLaVA-OneVision \cite{llavaov}, Vamba \cite{vamba}, Quicksviewer \cite{quicksviewer}, Video-XL \cite{videoxl}, LongVILA \cite{longvila}, LongVU \cite{longvu}, Video-R1 \cite{videor1}, Hour-LLaVA \cite{hourllava} and LongVITA \cite{longvita}, which are designed to perform video-level reasoning without external tool invocation. We also compare against multi-agent Video LLMs such as VideoMind \cite{videomind} and Video-RAG \cite{videorag}, which incorporate memory or retrieval mechanisms to aid in long-range reasoning. Finally, we evaluate against end-to-end agent Video LLMs, including Video-MTR \cite{videomtr}, Pixel Reasoner \cite{pixelreasoner}, VITAL \cite{vital} and FrameMind \cite{framethinker}.

\textbf{Benchmark.}
We evaluate our method on a set of comprehensive benchmarks designed to assess long video reasoning across various durations and task complexities. The benchmarks include MLVU \cite{mlvu}, VideoMME \cite{videomme} (without subtitles), LongVideoBench \cite{longvideobench}, LVBench \cite{lvbench} and MVBench \cite{mvbench}, each focusing on different aspects of reasoning over long-form video content. Across these benchmarks, the evaluation metric is accuracy (\%), providing a robust measure of model's video reasoning performance.

\begin{figure*}[t!]
   \centering
   \includegraphics[width=0.9\linewidth]{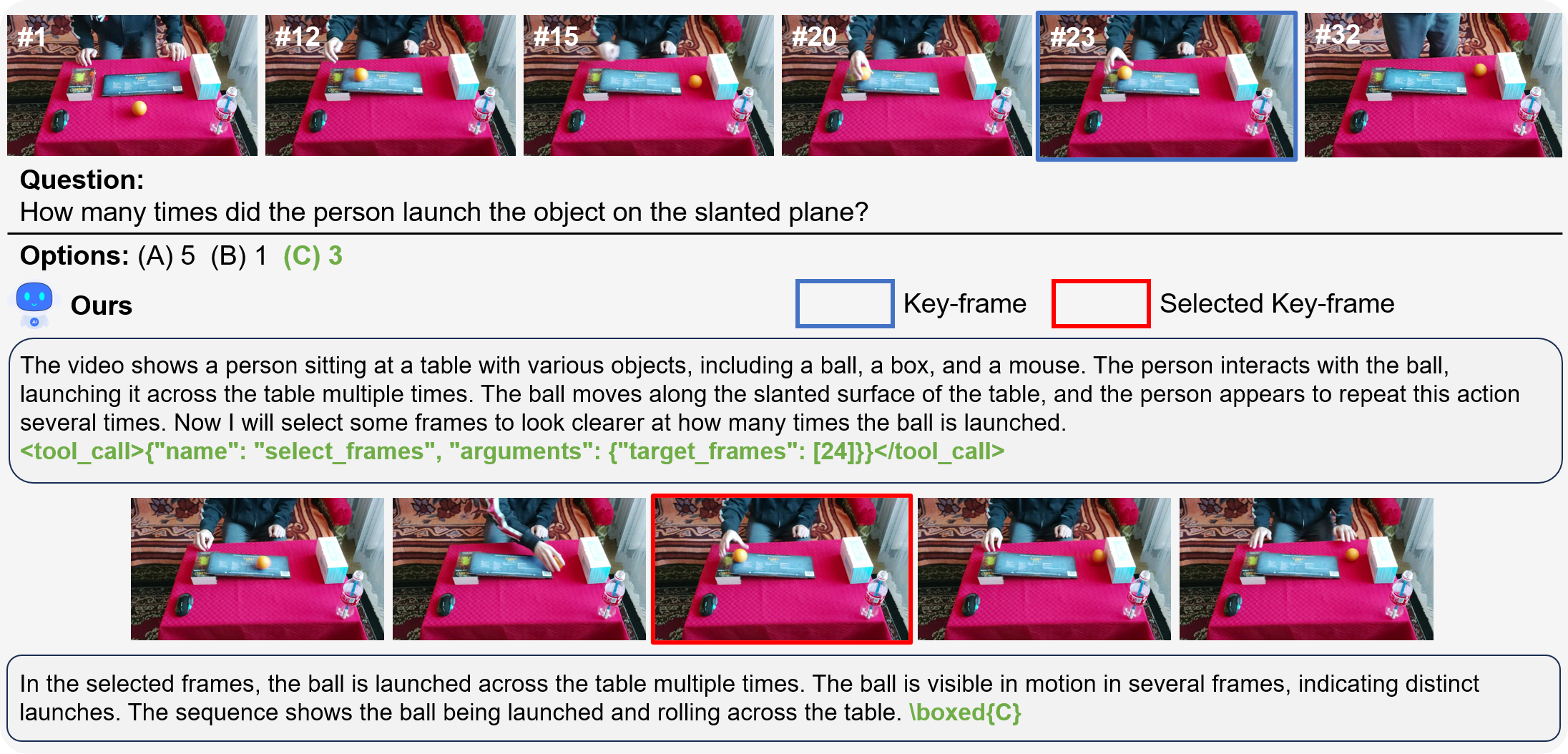}
   \vspace{-5pt}
   \caption{Our EARL framework ensures accuracy by actively interrogating evidence. For the complex counting question, the model uses a CoT step to identify the information gap, then calls the select\_frames tool for key-frame \#24. This action triggers localized re-sampling, providing a high-purity, fine-grained visual context. Utilizing this refined evidence stream, the model correctly answers.}
   \label{fig:case}
   \vspace{-15pt}
\end{figure*}

\subsection{Main Results}
To validate the effectiveness of our evidence-prioritized adaptive method, we perform a rigorous evaluation, with accuracy results detailed in Table \ref{tab:main_results}. We benchmark our performance against leading proprietary and open-sourced models.

Our evidence-prioritized adaptive method achieves superior performance among open-source models, significantly establishing a new state-of-the-art across all five demanding video reasoning benchmarks. Compared to other open-source Video MLLMs of comparable size and maximum frame capacity, our model consistently registers the highest accuracy. Notably, our method achieves 64.9\% on VideoMME \cite{videomme} (Overall) and 69.0\% on MVBench \cite{mvbench}, yielding a marked 11.3\% absolute improvement over the Qwen2.5-VL \cite{qwen25vl} baseline (53.6\% and 62.6\% respectively), which shares a similar foundation architecture. Furthermore, our results demonstrate robust superiority against existing agent-based approaches, as we significantly outperform all open-source end-to-end agent Video LLMs like Pixel Reasoner \cite{pixelreasoner} (67.8\% on MVBench) and FrameMind \cite{framethinker} (64.2\% on MVBench), which validates the necessity of our novel reward mechanism for enforcing evidence purity and reasoning accuracy.

The effectiveness of our method is particularly pronounced in long-video reasoning scenarios. The adaptive selection mechanism, combined with temporal refinement, allows our model to remain highly competitive with or even surpass many long-video models that rely on an order of magnitude larger fixed visual contexts. For instance, our model achieves a strong 57.8\% on VideoMME \cite{videomme} (Long), performing better than LongVA \cite{longva} (47.6\% with 128 frames) and LongVILA \cite{longvila} (53.0\% with 256 frames). This success validates that an intelligent, evidence-aware selection strategy is fundamentally more effective for high-quality reasoning than simply increasing the number of fixed input frames.

The superior performance stems from our strategy of precisely matching the visual context to the query's information needs. In long videos, a fixed, uniform sampling strategy inevitably includes many irrelevant frames, which dilute the limited context and hinder the Video LLMs' ability to focus on critical temporal cues. By actively discarding these redundant frames, our adaptive method provides the Video LLMs with a cleaner, high-density stream of relevant information. This targeted context delivery minimizes noise interference and maximizes the model's capacity for complex reasoning, which is essential for tackling the high-level semantic and temporal challenges present in these benchmarks. We provide a representative case in Figure \ref{fig:case} to show how our framework performs active evidence interrogation. For more cases, please refer to Appendix D.

\subsection{Ablation Study}
\subsubsection{Effectiveness of Evidence-Aware RL (EARL)}
\begin{figure}[htbp!]
  \centering
   \includegraphics[width=0.9\linewidth]{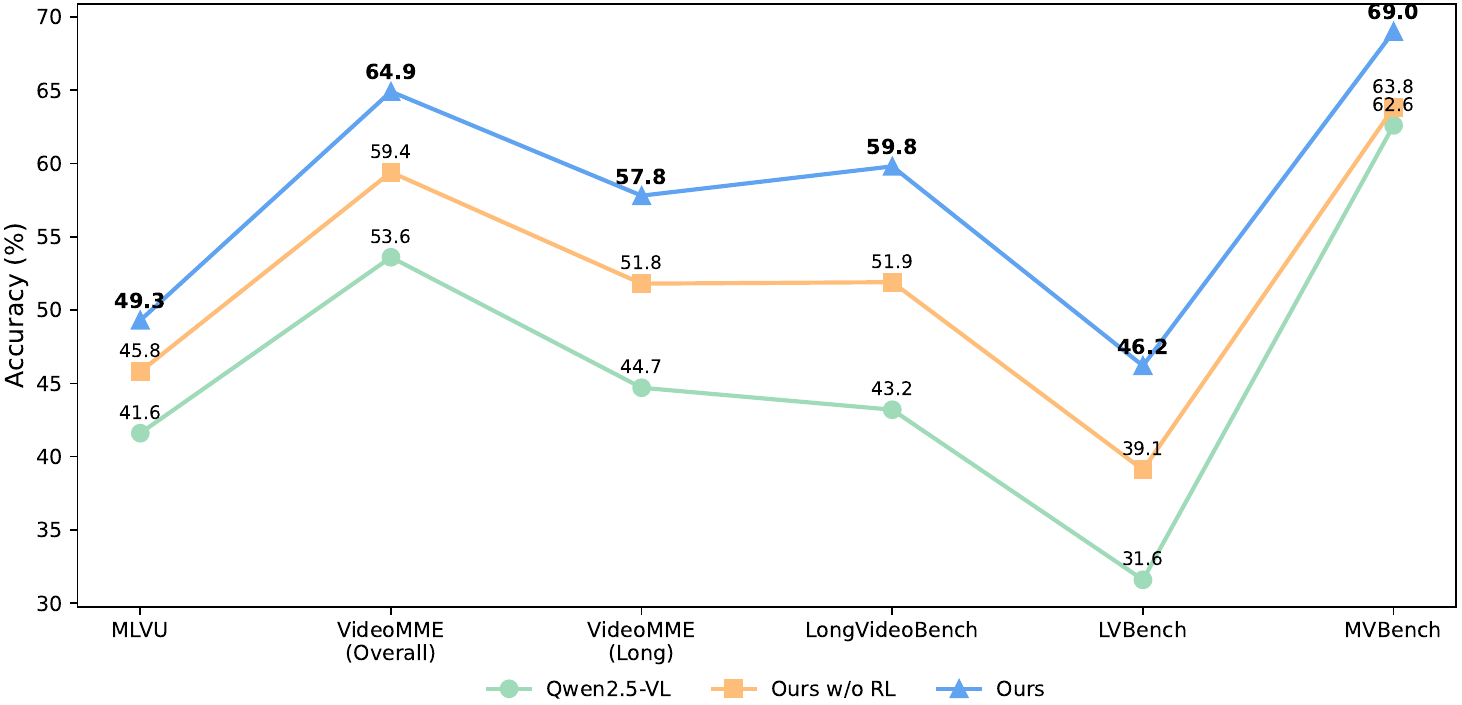}
   \vspace{-10pt}
   \caption{Ablation study on the effectiveness of EARL.}
   \label{fig:ablation_rl}
   \vspace{-15pt}
\end{figure}

The evidence-aware reinforcement learning (EARL) phase is paramount for transforming the basic operational competence learned in SFT into a precise and highly effective adaptive reasoning strategy. As shown in Figure \ref{fig:ablation_rl}, EARL is confirmed to be critical for achieving high accuracy through optimized visual input control. While SFT successfully enables the model to execute frame selection, the resulting imitation-based policy is suboptimal, leading to substantially lower accuracy. On the challenging LongVideoBench \cite{longvideobench}, EARL dramatically refines the strategy, boosting the SFT score from 51.9\% to 59.8\% (an absolute 7.9\% gain). Similarly, it increases accuracy on VideoMME \cite{videomme} (Long) from 51.8\% to 57.8\%, and on MVBench \cite{mvbench} from 63.8\% to 69.0\%. This consistent and significant improvement confirms the high effectiveness of the multi-component reward system. By overcoming the limitations of SFT's imitation learning, EARL successfully drives the model to select a highly informative set of frames, which directly translates into superior reasoning accuracy and demonstrates the full performance potential of our adaptive framework.

\subsubsection{Effectiveness of Relevance Reward}
\begin{figure}[htbp!]
    \vspace{-5pt}
   \centering
   \includegraphics[width=0.9\linewidth]{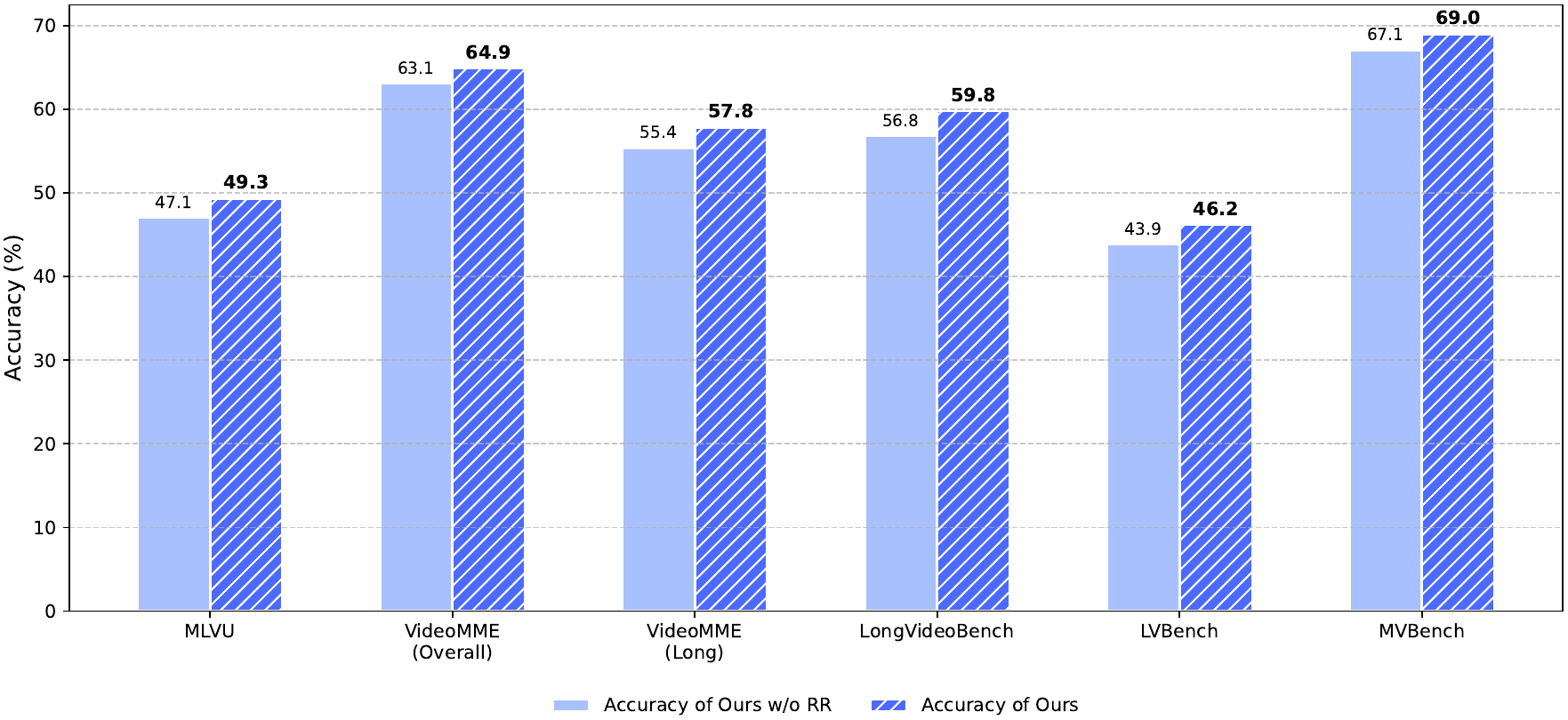}
   \vspace{-10pt}
   \caption{Ablation study on the effectiveness of relevance reward.}
   \label{fig:ablation_rr}
   \vspace{-15pt}
\end{figure}

We conduct an ablation study by removing the relevance reward component from the final training objective (Ours w/o RR in Figure \ref{fig:ablation_rr}). The results conclusively demonstrate that $r_{\text{relevance}}$ is indispensable for achieving high accuracy and controlling the visual context in adaptive frame selection. Without this reward, the model experiences significant degradation across all accuracy metrics. Removing $r_{\text{relevance}}$ directly harms performance. On LongVideoBench \cite{longvideobench}, accuracy drops from 59.8\% to 56.8\%, and on MLVU \cite{mlvu}, accuracy drops from 49.3\% to 47.1\%. This decline occurs because the excessive, irrelevant frames introduce noise and temporal distraction into the model's limited context window, thereby weakening the final reasoning accuracy. Thus, the relevance reward acts as a crucial filtering mechanism that enforces evidence purity and preserves the quality of the visual context.

\subsubsection{Effectiveness of IoU Constraint in Correct Reward}
The final component of our multi-part reward system is the IoU constraint embedded within $r_{\text{correct}}$. We perform an ablation study, Ours w/o IoU (Table \ref{tab:ablation_iou}), where the correctness reward is simplified to a standard binary reward, removing the requirement that selected frames must significantly overlap with the golden frames. This experiment is critical for demonstrating that the system must not only reward the correct final answer but also enforce reliance on a sequence of pure evidence—a pillar of a robust reasoning agent. The results clearly show that removing the IoU constraint degrades the overall accuracy, indicating a loss in the strategic quality of frame selection. On LongVideoBench \cite{longvideobench}, accuracy drops from the full model's 59.8\% to 57.8\%, and LVBench \cite{lvbench} sees a reduction from 46.2\% to 44.7\%. This phenomenon confirms that without the IoU constraint, the model is incentivized to find any path to the correct answer, even if that path involves selecting non-critical or suboptimal frames. Thus, the IoU constraint serves as a crucial supervisory signal during RL training, explicitly tying the output quality to the purity and relevance of the intermediate visual evidence.

\begin{table}[htbp!]
    \vspace{-5pt}
    \caption{Ablation study on the IoU constraint in correct reward.}
    \vspace{-5pt}
    \centering
    \resizebox{\linewidth}{!}{
        \begin{tabular}{l|cccccc}
        \toprule
        \textbf{Models} & \textbf{MLVU} & \multicolumn{2}{c}{\centering \textbf{VideoMME}} & \textbf{LongVideoBench} & \textbf{LVBench} & \textbf{MVBench} \\
        \midrule
        \shortstack{Ours w/o IoU}
        & \shortstack{47.9}
        & \shortstack{63.9}
        & \shortstack{56.4}
        & \shortstack{57.8}
        & \shortstack{44.7}
        & \shortstack{67.8}\\
        \cc \shortstack{\textbf{Ours}}
        & \cc \shortstack{\textbf{49.3}}
        & \cc \shortstack{\textbf{64.9}}
        & \cc \shortstack{\textbf{57.8}}
        & \cc \shortstack{\textbf{59.8}}
        & \cc \shortstack{\textbf{46.2}}
        & \cc \shortstack{\textbf{69.0}}\\
        \bottomrule
        \end{tabular}
    }
    \label{tab:ablation_iou}
    \vspace{-10pt}
\end{table}

\subsubsection{Effectiveness of the Dynamic Adjustment}
The final component we ablate is the dynamic adjustment (DA) mechanism, which controls the evolving balance between the accuracy reward and the relevance reward throughout training. By setting a fixed ratio $\alpha_\text{fixed}$ and $\beta_\text{fixed}$ (Ours w/o DA in Table \ref{tab:ablation_da}), we prevent the training from shifting its focus from initial strategy exploration to final policy refinement. The results show that the DA mechanism is crucial for achieving the model's final, highest-quality strategy. Without DA, accuracy consistently drops across all benchmarks, confirming that a statically balanced reward cannot guide the model to the optimal policy. We argue that the primary value of the DA mechanism lies in ensuring stable convergence to the best possible policy. By initially prioritizing the learning of correct answers and then gradually increasing the focus on pure frame selection, the DA mechanism prevents the early suppression of valuable exploration and guarantees that the policy is rigorously optimized for maximum accuracy and refined visual context.

\begin{table}[htbp!]
    \vspace{-5pt}
    \caption{Ablation study on dynamic adjustment.}
    \vspace{-5pt}
    \centering
    \resizebox{\linewidth}{!}{
        \begin{tabular}{l|cccccc}
        \toprule
        \textbf{Models} & \textbf{MLVU} & \multicolumn{2}{c}{\centering \textbf{VideoMME}} & \textbf{LongVideoBench} & \textbf{LVBench} & \textbf{MVBench} \\
        \midrule
        \shortstack{Ours w/o DA}
        & \shortstack{48.7}
        & \shortstack{64.6}
        & \shortstack{56.4}
        & \shortstack{58.4}
        & \shortstack{45.2}
        & \shortstack{68.3}\\
        \cc \shortstack{\textbf{Ours}}
        & \cc \shortstack{\textbf{49.3}}
        & \cc \shortstack{\textbf{64.9}}
        & \cc \shortstack{\textbf{57.8}}
        & \cc \shortstack{\textbf{59.8}}
        & \cc \shortstack{\textbf{46.2}}
        & \cc \shortstack{\textbf{69.0}}\\
        \bottomrule
        \end{tabular}
    }
    \label{tab:ablation_da}
    \vspace{-15pt}
\end{table}

\section{Conclusion}
\label{sec:conclusion}
In this work, we successfully addressed the critical challenges of visual redundancy and the lack of temporal granularity that plague long-form video reasoning in Video LLMs. Driven by our core philosophy, “Select Less, Reason More,” we introduced a novel framework for evidence-prioritized adaptive pixel-space video reasoning. Our central technical contribution is the evidence-aware reinforcement learning (EARL) framework, which transforms passive video processing into an active, strategic evidence interrogation process. We achieved this via two integrated innovations: a novel multi-component reward system designed to enforce evidence purity and reducing visual redundancy; and localized re-sampling around selected key frames to dynamically access the finer temporal detail for accurate decision-making. Rigorous evaluation across five demanding benchmarks confirms the superior performance of our EARL-trained model, establishing a new state-of-the-art among open-source Video LLMs. These results demonstrate that intelligent, frames-aware method is an effective and necessary direction for building scalable and high-performance Video LLMs.
\clearpage
{
    \small
    \bibliographystyle{ieeenat_fullname}
    \bibliography{main}
}

\end{document}